\documentclass[10pt,twocolumn,letterpaper]{article}

\usepackage{wacv}
\usepackage{times}
\usepackage{epsfig}
\usepackage{graphicx}
\usepackage{amsmath}
\usepackage{amssymb}
\usepackage{booktabs}
\usepackage{tabularx}
\usepackage{dsfont}
\usepackage{algorithm}
\usepackage{algpseudocode}
\usepackage{multirow}
\usepackage[export]{adjustbox}

\def\wacvPaperID{500} %

\wacvfinalcopy %

\ifwacvfinal
\def\assignedStartPage{9876} %
\fi

\newcommand{\mysubtitle}[1]{\noindent{\bf #1}}
\newcommand{\mysubsubtitle}[1]{\noindent{\bf \em #1}}

\ifwacvfinal
\usepackage[breaklinks=true,bookmarks=false]{hyperref}
\else
\usepackage[pagebackref=true,breaklinks=true,colorlinks,bookmarks=false]{hyperref}
\fi

\ifwacvfinal
\else
\fi

\begin{document}

\title{Localizing $\infty$-shaped fishes: Sketch-guided object localization in the wild}

\author{Pau Riba$^\dagger$ $^\ddagger$, Sounak Dey$^\dagger$ $^\ddagger$, Ali Furkan Biten$^\dagger$, Josep Lladós$^\dagger$ \\
$^\dagger$ Computer Vision Center, UAB, Spain\\
$^\ddagger$ Helsing AI, Germany\\
{\tt\small \{priba,sdey,abiten,josep\}@cvc.uab.cat}\\
{\tt\small \{pau.riba,sounak.dey\}@helsing.ai}
}

\maketitle

\begin{abstract}

This work investigates the problem of sketch-guided object localization (SGOL), where human sketches are used as queries to conduct the object localization in natural images. In this cross-modal setting, we first contribute with a tough-to-beat baseline that without any specific SGOL training is able to outperform the previous works 
on a fixed set of classes. The baseline is useful to analyze the performance of SGOL approaches based on available simple yet powerful methods. We advance prior arts by proposing a sketch-conditioned DETR (DEtection TRansformer) architecture which 
avoids a hard classification and alleviates the domain gap between sketches and images to localize object instances. %
Although the main goal of SGOL is focused on object detection, we explored its natural extension to sketch-guided instance segmentation. This novel task allows to move towards identifying the objects at pixel level, which is of key importance in several applications. We experimentally demonstrate that our model and its variants significantly advance over previous state-of-the-art results. All training and testing code of our model will be released to facilitate future research\footnote{\url{https://github.com/priba/sgol_wild}}.
\end{abstract}

\section{Introduction}\label{s:intro}
\begin{figure}
    \centering
    \includegraphics[width=\columnwidth]{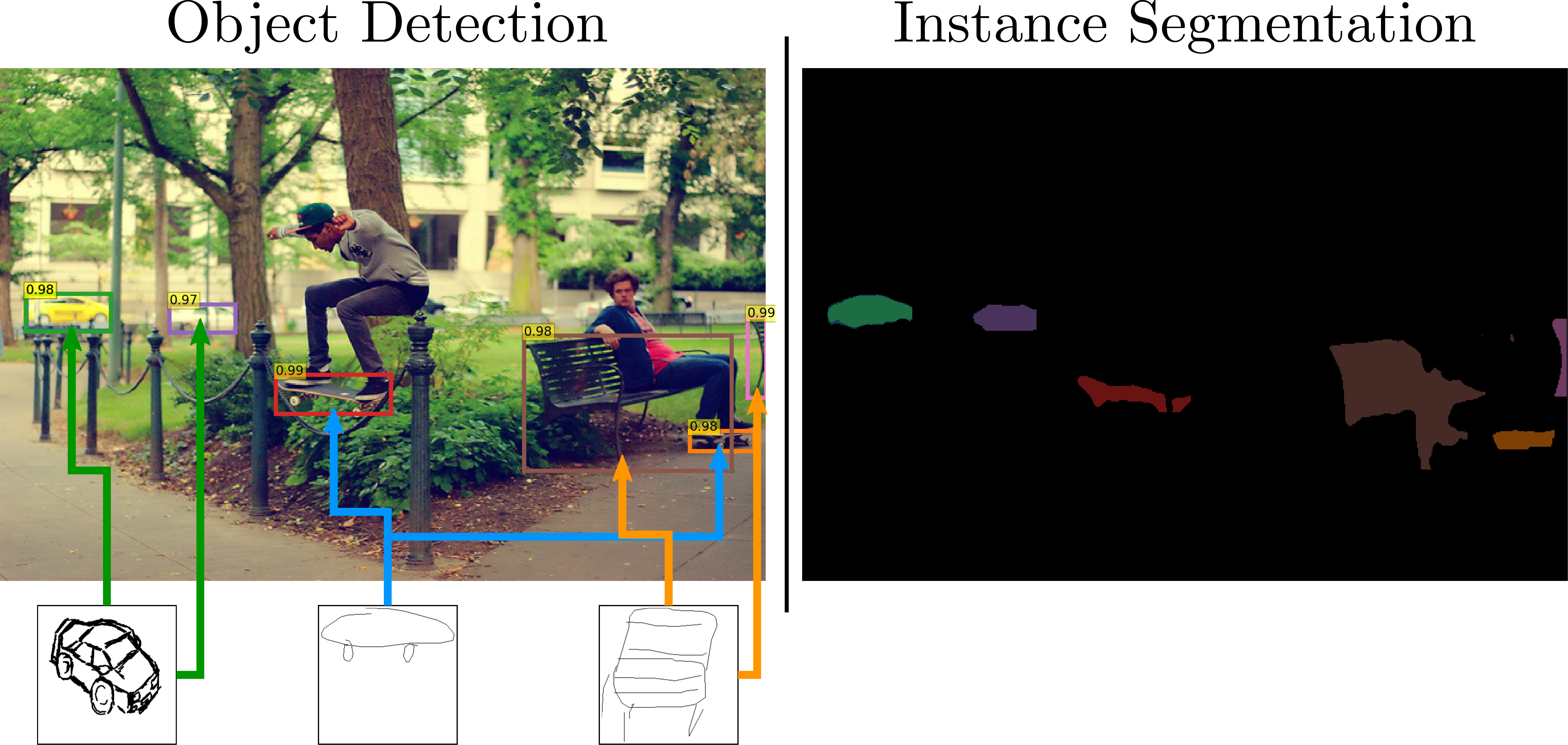}
    \caption{The sketch-guided object localization problem consists of two proposed tasks, SGOL at object detection (left) and instance segmentation (right) level. Three different sketches from two different datasets are used as queries to the same natural image. It is worth noting the emerging challenges,~\ie~abstraction level, position, occlusion, etc. (best viewed in color).}
    \label{f:teaser}
\end{figure}

Sketches have been used over centuries to convey information even before the invention of writing. By means of few strokes, humans have been able to create abstractions to communicate from difficult ideas to creative notions. Humans are well equipped to decipher this condensed information even at the absence of visual cues such as color or texture that are present in natural images. In addition, the variability on this modality is vast which can be found from the sketches of amateur to expert drawers or even from realistic to stylistic or artistic representations. Overall, containing condensed information with ample variability along with the pervasive nature of touchscreen and digital pen devices has motivated the emergence of the sketch modality for a broad range of tasks.

Out of all the tasks concerning sketch modality, sketch-based image retrieval (SBIR) has recently enjoyed a great deal of attention~\cite{bhunia2020sketch,dey2019zssbir,yu2017sketch,zhang2016sketchnet}. %
These frameworks have represented an important step towards the practical use of sketches in a real scenario. Even though SBIR requires some level of understanding of the image and the sketch, the task does not require the concept of `object' to be localized in an image. Natural evolution of SBIR dictates a more conclusive matching at the image level given the necessity of a better understanding of the sketch concepts. In other words, to face more ambitious challenges, the next step of SBIR is the localization of objects in an image given a sketch, namely, sketch-guided object localization (SGOL). Figure~\ref{f:teaser} visually illustrates this problem localizing three different sketches in the same natural image. Tripathi~\etal~\cite{tripathi2020sketch} proposed the first work to tackle such problem in terms of one-shot object detection given sketch as a query.

The problem of sketch-guided object localization is complex. It shares several challenges laid out in SBIR. In particular, there is a large domain gap between the modalities of sketches and natural images. This domain gap includes: (i) the lack of visual cues such as colors, texture, or context,~\ie~background; (ii) the use of canonical sketches as iconic representations of objects,~\ie~humans tend to abstract some instances using the same shape (imagine an $\infty$-shaped fish), which responds to a cultural convention of a concept more than an accurate visual representation; (iii) partial drawings, e.g.~to sketch only the face of a dog rather than the whole animal; and (iv) the high variability found in human sketches as a result of varying drawing skills and visual interpretations. 

Therefore, closing the domain gap at the pixel level (\ie aligning visual features between domains) is quite challenging. Instead, we propose to tackle the problem at semantic level. This leads us to the first contribution of this work, a tough-to-beat (T2B) baseline to close the domain gap at the semantic level. Specifically, since the sketches and images are paired at the level of semantic category, we utilize a pretrained object detector with a pretrained sketch classifier to match their labels and circumvent the domain gap.
Notably, we show that our tough-to-beat baseline surpasses the current state-of-the-art~\cite{tripathi2020sketch} while providing a good understanding of the task.
This kind of baselines is often neglected in favor of big and complicated models. We believe that our suggested baseline is critical in correctly assessing the future work's performance in SGOL. Since any advancement in an object detector can be reflected as an improvement of the task, the tough-to-beat baseline can characterize whether the boost in the performance are coming from the proposed method or the advances in the object detector. However, this baseline is only capable of dealing with classes provided at training time. This brings us to our second contribution, in which we propose a sketch-conditioned DETR architecture, referred as Sketch-DETR, that enables its usage in these cases where we are not restricted in a specific set of classes. Although the results show that our models outperform the current state-of-the-art~\cite{tripathi2020sketch} by a significant margin, Sketch-DETR is on par with our tough-to-beat baseline.

Furthermore, we also explore %
the obvious extension of a sketch-guided object localization at \emph{instance segmentation} level. In this setting, rather than simply providing a bounding box, we identify which pixels belong to the image object that matches to the query sketch. SGOL at instance segmentation level is of paramount interest in applications such as photo editing where a user can search an object given a sketch and then edit it (replace, delete, etc.).

Finally, 
the end goal of SGOL is to create models that perform well in open world %
scenarios. In other words, 
we would like to have models that can manage the unseen sketch classes. This formulation has been proposed by~\cite{tripathi2020sketch} where they divide the object classes into seen and unseen. We believe that before moving towards open world scenarios as intermediate step we have to deal with cross-dataset evaluation on sketches
Thus, we suggest to first solve the cross-dataset problem within sketches before progressing to the harder problems. To this end, we train our models on two common sketch datasets, Sketchy~\cite{sketchy2016sketchy} and QuickDraw~\cite{jonas2016quickdraw} and we put forward a cross-dataset evaluation where the task is to train on one dataset while the model is evaluated on the other. It can be observed that this setting includes several challenges already mentioned, namely the former is to close the large domain gap in terms of the different drawing skills while the latter is to deal with unseen sketch classes.

To summarize, the main contributions of this work are:
\begin{itemize} 
    
    \item We introduce a tough-to-beat (T2B) baseline for the two tasks of SGOL,~\ie~detection and segmentation. 
    The baseline can identify whether the improvements are coming from the proposed method or the advances in the object detector backbone.
    
    \item We present a variant of DETR~\cite{carion2020detr}, Sketch-DETR, for both SGOL tasks.
    
    \item We offer a cross-dataset evaluation setting that is the first necessary step towards an open world scenario.

    \item We explore %
    the next stride in SGOL, which is sketch-guided instance segmentation with strong baselines.
    
\end{itemize}

\section{Related Work}\label{s:soa}
\mysubtitle{Object Detection:}
Object detection aims to detect objects of predefined categories in an image.  State-of-the-art object detectors, has been mostly dominated by anchor-based detectors, which can be broadly divided into one-stage methods~\cite{lin2017focal, liu2016ssd}, making predictions w.r.t. anchors and two-stage methods~\cite{dai2016r, ren2016faster}, making prediction w.r.t. proposals. Most of them unanimously adopt variants of deep convolutional neural networks~\cite{he2016deep} as their backbones. Recent attention has been shifted towards anchor free detectors comprising of keypoint-based methods~\cite{law2018cornernet, zhou2019bottom} and center-based methods~\cite{kong2020foveabox, tian2019fcos}. Zhang~\etal~\cite{zhang2020bridging} demonstrate that the final performance of these systems heavily depends on the exact way these anchors are set. To circumvent the above stages and streamline the detection process for directly predicting the absolute boxes, Carion~\etal~\cite{carion2020detr} proposed object detection as a direct set prediction problem~\cite{redmon2016you} by adopting an encoder-decoder architecture based on transformers~\cite{vaswani2017transformers}.

\mysubtitle{Instance Segmentation:}
Instance segmentation~\cite{ dai2016instance, hariharan2014simultaneous} is a category-independent object detector generating proposals along with labeled pixels corresponding to each instance. This makes it a hybrid of semantic segmentation~\cite{long2015fully} and object detection where it is required to predict per-pixel segmentation mask for object instances. Most of the state-of-the-art methods~\cite{du2020spinenet, qiao2020detectors, zhang2020resnest} widely use Mask-RCNN~\cite{he2017mask} as a framework. Most modern methods employ a two-stage process~\cite{arora2016simple, he2017mask, liu2018path}, where the object proposals are first created and then foreground background segmentation is done for each bounding box. This task can be widely divided into proposal-based~\cite{huang2019mask, kirillov2019panoptic} and proposal-free~\cite{gao2019ssap, neven2019instance} methods. 

The goal of conditioned (also known as one-shot) object detection and instance segmentation is to develop models that can localize~\cite{hsieh2019one} and segment~\cite{michaelis2018one} objects from arbitrary categories when provided with a single visual example from that category. Gao~\etal~\cite{gao2019ssap} propose a one shot instance segmentation network that outputs a pixel pair affinity pyramid to compute whether two pixels belong to the same instance, they then combine this with a predicted semantic segmentation to output a single instance segmentation map. A range of one-shot segmentation task exist, for semantic segmentation~\cite{michaelis2018oneclutter, shaban2017one}, medical image segmentation~\cite{zhao2019data}, video segmentation~\cite{caelles2017one}, and more recently co-segmentation~\cite{hsu2019deepco}. 

\mysubtitle{Sketch-Image Multi-Modality:}
Free-hand sketch has several cross-modal applications when paired with other data modalities. 
Sketches can be used to retrieve natural photos~\cite{eitz2012hdhso, hu2013performance, liu2017deep, yu2017sketch, zhang2018generative}, manga~\cite{matsui2015challenge}, 3D shape~\cite{wang2015sketch}, video~\cite{rossetto2014cineast} and many more. Specifically, sketch based image retrieval (SBIR)~\cite{dey2018learning} has seen lot of research done to bridge this domain gap between sparse line drawings with dense pixel representations using cross modal deep learning techniques~\cite{dey2018aligning}. In these lines, different work has also been conducted on fine-grained SBIR~\cite{yu2016sketch}, zero-shot SBIR~\cite{dey2019zssbir}, large-scale SBIR~\cite{shen2018zero}, and many others. 

Recently, the important work~\cite{tripathi2020sketch} on SGOL uses a sketch query to localize all instances of an object in an image that also contains many other objects in a cross-domain scenario. 
It also addresses the task of detecting multiple objects in the same image using novel feature and attention fusion strategies.
As well, providing plausible baselines with modified Faster R-CNN~\cite{ren2016faster}. The work also tackles the problem of zero-shot in SGOL, where given sketches of an unseen category the model tries and localizes object instances of that category present in the image.
Finally, Hu~\etal~\cite{hu2020sketch} proposed the first work with the goal of segment a query sketch in a natural image. In particular they deal with the fine-grained and category-level segmentation.

Our work is placed in a category-level SGOL that similarly to the work of~\cite{tripathi2020sketch}, is able to localize a query sketch in a natural image. In addition, following~\cite{hu2020sketch}, we also extend this work to instance level segmentation  by identifying the pixels belonging to the query sketch. Moreover, we put forward a tough-to-beat baseline which outperforms previous state-of-the-art.

\section{Method}\label{s:sgol}
\subsection{Problem formulation} 

Let \(\mathcal{C}\) be the set of all possible categories that are common for an image and a sketch datasets. Considering \(\{\mathcal{X}, \mathcal{B}, \mathcal{Y}\}\) to be an object detection or instance segmentation dataset, which contains natural images \(\mathcal{X}=\{x_i\}_{i=1}^N\), and its corresponding set of annotated object instances \(\mathcal{B} = \{B_i\}_{i=1}^N\)  where \(B_i = \{b_{ij}\}_{j=1}^{n_i}\) along with their corresponding category \(\mathcal{Y} = \{Y_i\}_{i=1}^N\) where \(Y_i = \{y_{ij} \in \mathcal{C}\}_{j=1}^{n_i}\). In addition, let \(\{\mathcal{S}, \mathcal{Z}\}\) be a sketch dataset containing the set of sketch images \(\mathcal{S}=\{s_i\}_{i=1}^M\) and their corresponding categories \(\mathcal{Z} = \{z_i \in \mathcal{C}\}_{i=1}^M\). Now, given a natural image \(x_i \in \mathcal{X}\) with ground-truth \(B_i \in \mathcal{B}\) and \(Y_i \in \mathcal{Y}\) on the one hand, and given a query sketch \(s\in \mathcal{S}\) belonging to the category \(z \in \mathcal{C}\) on the other hand; the proposed object detection or instance segmentation model \(\phi(\cdot \mid \cdot)\) has the ability to properly obtain the corresponding set of \(K\) instances \(\phi(x_i \mid s) = \Tilde{B_i} = \{\Tilde{b}_{ij}\}_{j=1}^K \subseteq B\) such that its corresponding set of categories are all the same, \ie \(y_{ij} = z\), as the sketch category \(z \in \mathcal{C}\).

\mysubtitle{Cross-dataset formulation:} 
The formulation of disjoint setting from~\cite{tripathi2020sketch} (disjoint IID in Table~\ref{tab:eval}) does not resemble the real application scenario compared to disjoint OOD since there is always a domain gap in real world. Moreover, we follow the SBIR literature~\cite{dey2019zssbir, Pang_2020_CVPR} on dealing with the domain gap before moving into the harder task of disjoint setting. Thus, we put forward a cross-dataset evaluation on sketches as an intermediate stage towards the desired disjoint framework. 

\begin{table}[tb]
    \centering
    \caption{Our proposed evaluation divides the problem between common and disjoint train-test categories and between same dataset and cross dataset settings.}
\begin{tabular}{ccc}
    \hline
     \textbf{Evaluation}  & \textbf{Common}  & \textbf{Disjoint} \\
     \hline
    \textbf{Same-Dataset} & Constrained~\cite{tripathi2020sketch} & Disjoint IID~\cite{tripathi2020sketch}  \\
  \textbf{Cross-Dataset} & Intermediate Stage & Disjoint OOD\\
     \hline
    \end{tabular}
    \label{tab:eval}
\end{table}

This challenging cross-dataset problem is formally defined as follows. Let \(\{\mathcal{X}, \mathcal{B}, \mathcal{Y}\}\), \(\{\mathcal{S}^{\mathrm{tr}}, \mathcal{Z}^{\mathrm{tr}}\}\) and, \(\{\mathcal{S}^{\mathrm{te}}, \mathcal{Z}^{\mathrm{te}}\}\) be an object detection/instance segmentation dataset, a training sketch dataset and a testing sketch dataset, respectively. Then, given a trained  model \(\phi(\cdot \mid \cdot)\) on the corresponding natural images \(\mathcal{X}\) and sketches \(\mathcal{S}^{\mathrm{tr}}\), the evaluation is performed on an unseen data \(\mathcal{S}^{\mathrm{te}}\). Note that in this setting, a given model has to deal with two important challenges: (i) the large difference in level of abstraction in two datasets; and (ii) novel classes between datasets,~\ie~ \(\mathcal{C}^{\mathrm{tr}} \cap \mathcal{C}^{\mathrm{te}} \neq \emptyset\) and \(\mathcal{C}^{\mathrm{te}} \not\subseteq \mathcal{C}^{\mathrm{tr}}\). Thus, this setting evaluates the ability of our model \(\phi(\cdot \mid \cdot)\) to extrapolate to new drawing skills as well as to unseen categories without requiring further training.

Taking into account both settings, two models have been proposed. Firstly, a tough-to-beat baseline for SGOL which is able to outperform any state-of-the-art method in the constrained setting (see Table~\ref{tab:eval}). Secondly, a conditioned DETR model that is able to deal with an open world setting.

\subsection{Tough-to-beat baseline for SGOL}

\begin{figure*}[t]
    \centering
    \includegraphics[width=\textwidth]{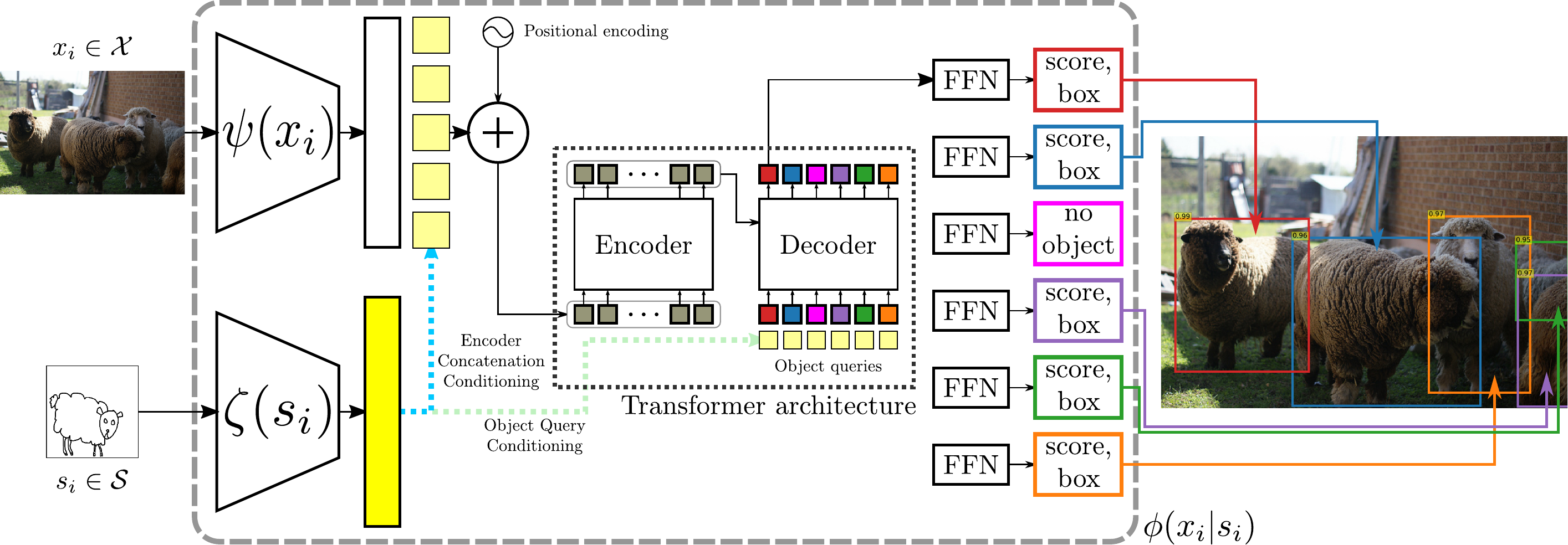}
    \caption{Overview of the sketch conditioned DETR. Our model \(\phi(\cdot \mid \cdot)\) accept a given natural image \(x_i\in \mathcal{X}\) and a query sketch \(s_i \in \mathcal{S}\). Firstly, the two CNN backbones  \(\psi(\cdot)\) and \(\zeta(\cdot)\) outputs a feature map for each input respectively. Secondly, these feature maps are merged and flattened just before feeding them into an encoder-decoder transformer architecture. Finally, a shared feed forward network (FFN) is used to obtain the final bounding box and score prediction (optionally, the segmentation masks are also predicted). Note the optional object query conditioning (in \textcolor{green}{green} dotted lines) and the encoder concatenation conditioning (in \textcolor{blue}{blue} dotted lines).}
    \label{f:detr_overview}
\end{figure*}

In general, 
big and complicated methods are designed to face novel problems.
Inspired by the work proposed by Arora~\etal~\cite{arora2016simple}, we realize that previous arts on SGOL often neglect simple yet powerful baselines. In this section, we describe our contribution consisting in such T2B baselines by exploiting state-of-the-art pretrained object detectors (same for instance segmentation approach) and trained sketch classifiers. From this point on, let us consider \(\psi(\cdot)\) and \(\zeta(\cdot)\) such pretrained models.

\begin{algorithm}[b]
\small
{
\hspace*{\algorithmicindent} \textbf{Input: } Test natural image and sketch test sets \(\{\mathcal{X}_T,\mathcal{S}_T\}\);\\ \hspace*{\algorithmicindent}\hspace*{\algorithmicindent} Pretrained weights \(\{\Theta_{\psi}, \Theta_{\zeta}\}\) \\ 
\hspace*{\algorithmicindent} \textbf{Output: } Detection set \(\{\Tilde{B_i}\}_{i=1}^N\)
\begin{algorithmic}[1]
\For{\(\{x_i, s_i\} \in \{\mathcal{X}_T, \mathcal{S}_T\}\)}
\State \(\bar{B}_i, \bar{Y}_i \leftarrow \psi(x_i; \Theta_{\psi})\)
\State \(\bar{z}_i \leftarrow \zeta(s_i)\)
\State \(\Tilde{B}_i \leftarrow \{ \bar{b}_{ij} \in \bar{B}_i ; \mathds{1}_{\bar{y}_{ij} = \bar{z}_i}\text{,  where }\bar{y}_{ij}\in \bar{Y}_i \}\)
\EndFor
\end{algorithmic}
}
\caption{Inference algorithm for the Tough-to-beat baseline given a test set containing pairs of natural images and sketches \(\{x_i, s_i\}\in \{\mathcal{X}_T,\mathcal{S}_T\}\).} \label{alg:tough}
\end{algorithm}

Our T2B method \(\phi(\cdot \mid \cdot)\) for the task of SGOL, obtains the objects proposals for the particular image, with any pretrained object detection or instance segmentation model. These object proposals are then filtered according to the class of the sketch given by a pretrained sketch classifier, hence, bypassing any specific joint training on the problem of sketch-guided object detection. An independent training for detecting objects on natural images and classifying sketches to their corresponding classes is adequate. It must be noted that this setting requires an intersection between the categories of the objects and sketches. 
Thus, given a natural image \(x_i\in\mathcal{X}_T\) from test set and the conditioning sketch \(s \in \mathcal{S}_T\), we can define our model as \(\phi(x_i \mid s) = \{ \bar{b}_{ij} \in \bar{B}_i ; \mathds{1}_{\bar{y}_{ij} = \bar{z}}\text{,  where } \bar{B}_i, \bar{Y}_i = \psi(x_i; \Theta_{\psi})\text{ and } \bar{y}_{ij}\in \bar{Y}_i \}\). A detailed overview of this tough-to-beat baseline is introduced in Algorithm~\ref{alg:tough}.

Although our proposed T2B method has shown to be formidable, which is able to outperform specific and complicated SGOL methods, it has a limitation that makes it incapable to deal with unconstrained scenarios. Our T2B baseline depends on pretrained classifiers, for both the sketches and detected objects that have been trained with a closed set of classes. Thus, given a query from unseen class it fails. To circumvent this drawback, in the following section, we propose a novel approach which conditions the DETR object detection model on a given sketch image.

\subsection{Sketch Conditioned DETR}

Based on the successful work proposed by Carion~\etal~\cite{carion2020detr} named DETR (DEtection TRansformer) we propose a variant, namely Sketch-DETR, to tackle the problem of SGOL.

\mysubtitle{DETR architecture~\cite{carion2020detr}:} The main components of DETR architecture are a CNN backbone, an encoder-decoder transformer and feed forward networks (FFNs) to obtain the final predictions. In addition, a set-based bipartite matching loss forces a unique prediction for each ground-truth bounding box. Its main feature outperforming any other object detector is its capacity for reasoning about the relations among the different objects provided by the self-attention mechanism of the transformer architecture. Figure~\ref{f:detr_overview} shows an overview of the proposed architecture.

The DETR model consists of a CNN backbone (ResNet-50~\cite{he2016deep}) which extracts the input feature maps \(f \in \mathbb{R}^{d \times H \times W}\). These features are then flattened to a sequence \(z \in \mathbb{R}^{d \times HW}\) suitable for the traditional transformer encoder. 
This flattening operation removes the spatial information of features. Therefore, to avoid the permutation-invariant characteristics of the transformer-based architecture, this model requires a positional encoder to be jointly fed.
Later, as we can see in Figure~\ref{f:detr_overview}, the decoder transformer considers the previously processed features from the encoder on the one hand, and a set of \(N\) learnable embeddings of size \(d\) denoted as object queries \(q = \{q_i \in \mathbb{R}^d\}_{i=1}^N\). These embeddings, randomly initialized, allow the transformer to localize \(N\) objects in parallel. Finally, each output of the decoder is processed by a prediction feed-forward network heads which predicts the bounding box in terms of the normalized center coordinates, height and width, and the class of each predicted object. Note that not every object query corresponds to an object, therefore, a special empty class with label \(\emptyset\) is used to represent the no object label. Moreover, DETR can be extended by adding a mask head on top of the decoder outputs.

\mysubtitle{Sketch-DETR:} Given a sketch \(s\in \mathcal{S}\), our model proposes two different conditioning methods for the DETR detector. First, we explore to incorporate the sketch condition at the object query level. Second, the sketch features are introduced jointly at the transformer encoder. Let us define \(\zeta(\cdot)\) as a CNN backbone to extract sketch features \(f_s = \zeta(s) \in \mathbb{R}^d\).

\mysubsubtitle{Object Query Conditioning:} Taking into account the DETR formulation, a natural way of conditioning its output to a given sketch features \(f_s\) is to feed these features to the transformer decoder jointly with the query objects. The main motivation of this conditioning scheme is to provide the query object not only with spatial information but also about the contents it should search. Thus, we process with a simple linear layer the concatenation of the query object \(q_i\) and the sketch features \(f_s\) in order to get a new tensor \(\Tilde{q_i} = [q_i; f_s]\) of shape of \(d\) dimensions. 

\mysubsubtitle{Encoder Concatenation Conditioning:} The second conditioning method combines the features \(f \in \mathbb{R}^{d\times H \times W}\) obtained by the DETR CNN backbone with the features \(f_s \in \mathbb{R}^d\) obtained by the sketch CNN backbone by means of a concatenation. Note that \(f_s\) is repeated accordingly to match the size of \(f\). Afterwards, a \(1\times 1\) convolution reduces its dimensions to \(d\times H \times W\). The motivation of such conditioning is to enhance the features of these regions with high correlation with the given sketch.

Figure~\ref{f:detr_overview} describes both conditioning methods with a green and blue dotted arrow, corresponding to object query and encoder concatenation.

\mysubtitle{Learning objectives:} Following the same scheme proposed by Carion~\etal~\cite{carion2020detr}, we have tweaked their learning objectives for the binary case, \ie~either it is the searched object or not. Firstly, we require to find a proper matching between the set of \(N\) predicted objects with the ground-truth set of objects. It can be acknowledged that the ground-truth will be padded with \(\emptyset\) (no object) to meet the number of predictions \(N\). In this regard, the authors propose a bipartite matching between these two sets to predict the proper permutation \(\Tilde{\sigma} \in \mathfrak{G}_N\) with the lowest cost,
\begin{equation}
    \Tilde{\sigma}_i=\underset{\tau \in \mathfrak{G}_{N}}{\operatorname{argmin}} \sum_{j=1}^{N} \mathcal{L}_{\operatorname{match}}\left(b_{ij}, \Tilde{b}_{i\tau(j)}\right)
\end{equation}
where \(\mathcal{L}_{\operatorname{match}}\) is a pair-wise matching cost between the ground-truth and the prediction. This cost considers the predicted box as well as the objectness score. Afterwards, taking into account the assigned pairs, the Hungarian loss is computed as follows,
\begin{equation}
\begin{split}
    \mathcal{L}_{\operatorname{Hungarian}}(b_{ij}, \Tilde{b}_{ij}) = &\sum_{i=1}^N  \left[-\log p_{\Tilde{\sigma}(i)}(c_i) + \right. \\
    & \left. \mathds{1}_{\{c_i\neq \emptyset\}} \mathcal{L}_{\operatorname{box}}(b_{ij},\Tilde{b}_{i\Tilde{\sigma}(j)}) \right],
\end{split}
\end{equation}
where \(p_{\Tilde{\sigma}(i)}(c_i)\) is the probability of class \(c_i\) for the prediction with index \(\Tilde{\sigma}(i)\), and \(\mathcal{L}_{\operatorname{box}}\) weights the Generalized Intersection over Union (IoU)~\cite{rezatofighi2019generalized} and an \(L_1\) loss among the predicted boxes. Thus, the final loss for the predicted boxes is defined as
\begin{equation}
\begin{split}
    \mathcal{L}_{\operatorname{box}}(b_{ij}, \hat{b}_{i\Tilde{\sigma}(j)}) = &\lambda_{\operatorname{iou}}\mathcal{L}_{\operatorname{iou}}(b_{ij}, \hat{b}_{i\Tilde{\sigma}(j)}) + \\ 
    & \quad \lambda_{\operatorname{L_1}} ||b_{ij} - \hat{b}_{i\Tilde{\sigma}(j)}||_1.
\end{split}
\end{equation}

Additionally, two supplementary losses are optionally used for the instance segmentation mask prediction. On the one hand, the Focal Loss~\cite{lin2017focal} which deals with the imbalance present in the data and, on the other hand the DICE/F-1 loss~\cite{milletari2016v} that directly optimizes the Dice coefficient given the predicted masks \(\Tilde{m}\),
\begin{equation}
    \mathcal{L}_{\operatorname{DICE}}(m, \Tilde{m}) = 1 - \frac{2 m \sigma(\Tilde{m}) + 1}{\sigma(\Tilde{m}) + m + 1},
\end{equation}
where \(\sigma(\cdot)\) is the sigmoid function.

\section{Experimental Validation}\label{s:experiment}

\begin{table*}[!htb]
    \centering
    \caption{Comparison against the state-of-the-art methods with that of the proposed model for the sketch-guided object detection task. Upper Bound assumes a perfect sketch classifier taking into account the Tough-To-Beat setting.}
    \begin{tabular}{l l c c c c c}
    \toprule
        \textbf{Model} & \textbf{Variant} & \multicolumn{2}{c}{\textbf{Sketchy}} && \multicolumn{2}{c}{\textbf{Quick-Draw}} \\
        \cmidrule{3-4} \cmidrule{6-7}
        & & \textbf{mAP} & \textbf{AP@50} && \textbf{mAP} & \textbf{AP@50}\\
        \midrule
        Modified Faster R-CNN~\cite{tripathi2020sketch} & - & - & - && 18.0 & 31.5 \\
        Matchnet~\cite{hsieh2019matchnet} & - & - & - && 28.0 & 48.5 \\
        Cross-modal attention~\cite{tripathi2020sketch} & - & - & - && 30.0 &  50.0 \\
        \midrule
        DETR Upper Bound & - & 48.0 & 69.9 && 42.8 & 65.9 \\
        \midrule
        \multirow{3}{*}{Tough-To-Beat} & Faster R-CNN~\cite{ren2016faster} & 40.2 & 64.4 && 35.5 & 58.1 \\
         & RetinaNet~\cite{lin2017focal} & 42.5 & 66.1 && 37.9 & 60.1 \\
         & DETR~\cite{carion2020detr} & \textbf{47.0} & \textbf{68.7} && 41.1 &  \textbf{62.7} \\
        \midrule
        \multirow{2}{*}{Sketch-DETR} & Object Query & 33.3 & 52.2 && 38.7 & 57.5 \\
        & Encoder Concatenation & 42.0 & 63.6 && \textbf{41.4} & 62.1 \\ %
    \bottomrule
    \end{tabular}
    \label{t:detection_comparison}
\end{table*}

This section experimentally validates the proposed SGOL approach on a real-life multi-object image dataset along with two other standard datasets of sketches. We present a detailed comparison with the state-of-the-art as well as a novel yet relevant cross-domain evaluation.

\subsection{Dataset and Implementation Details}

\mysubtitle{Sketchy Dataset~\cite{sketchy2016sketchy}:} Originally created as a fine-grained association between sketches to particular photos for fine-grained retrieval. The Sketchy dataset contains 75,471 hand-drawn sketches of 12,500 object photos belonging to 125 different categories.

\mysubtitle{QuickDraw Dataset~\cite{jonas2016quickdraw}:} It is a huge collection of drawings (50M) belonging to 345 categories obtained from the \emph{Quick, Draw!}\footnote{\url{https://quickdraw.withgoogle.com/}} game. In this game, the user is asked to draw a sketch of a given category while the computer tries to classify them. The way the sketches are collected provides the dataset a large variability, derived from human abstraction.

\mysubtitle{MSCOCO Dataset~\cite{lin2014microsoft}:} A large-scale image dataset which has been widely used in object detection and instance segmentation research. Including the background class it has a total of 81 classes,  with dense object bounding box annotations.

\mysubtitle{Implementation details:} Following the same protocol introduced in~\cite{carion2020detr}, we train our model with the AdamW~\cite{loshchilov2018decoupled} optimizer. We started the training from a COCO pretrained DETR model with ResNet-50 as a backbone ($\psi(\cdot)$). The same backbone has been used for the sketch encoder \(\zeta(\cdot)\) that has been trained as a classification model for the corresponding sketch dataset (with the COCO intersecting classes). In training phase, the weights for the backbones as well as for the transformer encoder have been frozen. The training takes 50 epochs for the object detection and 25 epochs for the instance segmentation, once we initialized the weights with pretrained DETR on object detection.
The whole framework was implemented with PyTorch~\cite{paszke2019pytorch}.

There are overlapping classes between MSCOCO and sketch datasets. With abuse of notation, these overlapping classes between MSCOCO and Quickdraw are referred as $Q$ and the overlapping classes between MSCOCO and Sketchy are referred as $S$.
In addition, $Q \cap S$ denotes the common classes between QuickDraw and Sketchy. $Q\setminus S$ corresponds to the classes that are in QuickDraw but not in the Sketchy and vice versa. The number of classes for $Q$ and $S$ are 56 and 23, respectively.

In our experiments, we use the whole Sketchy dataset on the one hand and, similarly to \cite{tripathi2020sketch}, we randomly selected a total of 800K sketches across these common classes for QuickDraw. To ensure the reproducibility of results at test time we use $20$ unseen sketches in training per category from Sketchy and QuickDraw dataset in coherence with the experimental setup of ~\cite{tripathi2020sketch}.
We trained and evaluated our model on MSCOCO-2017 dataset for SGOL on object detection and instance segmentation. 

\mysubtitle{Evaluation protocol:} The proposed evaluation provides the traditional object detection and instance segmentation metrics, similarly to the previous work of Tripathi~\etal~\cite{tripathi2020sketch}.
Thus, all the experiments show the \%AP@50 and \%mAP scores in accordance to~\cite{lin2014microsoft}.
Note that this evaluation does only consider detection from the desired class rather than possible visually similar detections that might be considered correct by human evaluator.

\subsection{Sketch-guided object localization}

\mysubtitle{SGOL - Object Detection:} Table~\ref{t:detection_comparison} provides comparisons of our models against the state-of-the-art.  We report a comparative study of methods presented in~\cite{tripathi2020sketch} for the application of SGOL, namely Modified Faster R-CNN, Matchnet~\cite{hsieh2019matchnet} along with their own method of Cross-modal attention. Even though the specific samples used in their implementation were not available, we have reproduced the same evaluation setting in terms of the used classes and the number of sketches and images. 

We demonstrate that even using the same object detector proposed by Tripathi~\etal~\cite{tripathi2020sketch}, namely Faster R-CNN~\cite{ren2016faster}, our T2B baseline is able to outperform the cross-modal attention by \(5\%\) and \(8\%\) in terms of mAP and AP@50. This exhibits two things, the former is that the capacity of their model in a closed scenario is not achieved. More importantly, the latter is that the improvement achieved by the their model can be surpassed by a simpler method. 

Apart from providing the figures for Faster RCNN~\cite{girshick2015fast}, we present RetinaNet~\cite{lin2017focal} to see the effect of object detectors' performance. As can be appreciated from the table, the performance of the object detector is preserved on the SGOL task. 
We further calculate the upper bound of T2B baseline DETR, which assumes a perfect sketch classifier, showing that the performance is mainly bounded by the performance of the used object detection as the sketch classifier performs with a high accuracy. The idea of the upper bound is to display the comprehensive capacity of the aforementioned model.  

Regarding the variant of Sketch-DETR, we demonstrate that concatenating sketch features at the encoder level instead of object query level significantly boosts the numbers. Our conjecture is that conditioning at the object queries requires a higher level of abstraction than providing the sketch features at the encoder level. In this regard, the object query conditioning completely relies on the self-attention mechanism. Concerning the scores in Sketchy dataset, the baseline shows a better performance. However, our Sketch-DETR model obtains a slightly drop on the performance. We hypothesize that the size of the dataset is too small to take advantage of the whole potential of the DETR model.

Moreover, even though we obtain 11\% improvement in mAP and 13\% in AP@50 in the QuickDraw dataset over~\cite{tripathi2020sketch} with Sketch-DETR, as pointed out already, the improvement can not be directly compared to previous state-of-the-art because of the different backbones. Instead, comparing Sketch-DETR with T2B is more meaningful. Accordingly, we notice that both models achieve the same performance where the equivalent behavior can not be observed on Cross-Modal Attention~\cite{tripathi2020sketch}. We would like to point out although Sketch-DETR and T2B-DETR perform similarly, the advantage of Sketch-DETR is better monitored on open world scenario, in which T2B-DETR is not suited for.

We firmly believe that SGOL as a new sub-field has a lot of promise, however we want to emphasize that to correctly evaluate our models' advances, it is imperative to provide the tough-to-beat baseline as a starting point for any future work. This will allow the community to identify if the increase is coming from the advances in object detection or from the architecture proposed for SGOL.

\mysubtitle{SGOL - Instance Segmentation:} In Table~\ref{t:segmentation_comparison}, we show the study of our proposed model for the novel problem of SGOL on instance segmentation. 
It can be acknowledged that the results of instance segmentation is not on par with T2B as in the case of object detection. We believe that the task of delineating each distinct object of interest appearing in an image at pixel level is far harder task than the object detection. From an in-depth study of these results, we hypothesize that the sparse nature of some masks, especially on occluded objects, generates masks with low confidence scores and, therefore, not properly defined for a fine-grained segmentation but a coarse one.

Moreover, the proposed baselines is a guideline that methods should follow to understand exactly where the models lag behind. This is especially important when we try to assess our model's limitations. In that sense, our T2B approach provides the expected performance of a conditioned instance segmentation which is able to properly reason among the image and sketch contents with similar capacity of the backbone instance segmentation method.

\begin{table}[!htb]
    \centering
    \caption{Comparison against the state-of-the-art methods with that of the proposed model for SGOL at instance segmentation level.}
    \begin{tabular}{l c c c c c}
    \toprule
        \textbf{Model} & \multicolumn{2}{c}{Sketchy} && \multicolumn{2}{c}{Quick-Draw} \\
        \cmidrule{2-3} \cmidrule{5-6}
        & \textbf{mAP} & \textbf{AP@50} && \textbf{mAP} & \textbf{AP@50}\\
        \midrule
        DETR UB & 33.5 & 60.9 && 33.2 & 59.8 \\
        T2B-DETR & 32.8 & 59.7 && 30.2 & 54.6\\
        \midrule
        Sketch-DETR & 24.3 & 46.5 && 22.8 & 44.2 \\
    \bottomrule
    \end{tabular}
    \label{t:segmentation_comparison}
\end{table}

\begin{table}[!htb]
    \centering
    \caption{Cross-dataset evaluation among QuickDraw and Sketchy datasets. Training is performed in one sketch dataset whereas evaluation is performed in the second one. Note that we separate the seen \(Q\cap S\) and unseen classes \(Q \setminus S\).}
    \begin{tabular}{l l l c c}
    \toprule
        \textbf{Model} & \textbf{Train} & \textbf{Evaluated} &  \textbf{mAP} & \textbf{AP@50} \\
        \midrule
        T2B-DETR & SK & QD (\(Q \cap S\)) & 8.6 & 12.6  \\
        S-DETR & SK & QD (\(Q \cap S\)) & 37.6 & 55.9  \\
        S-DETR & SK & QD (\(Q \setminus S\)) & 1.5 & 2.8 \\ %
        \midrule
        T2B-DETR & QD & SK (\(Q \cap S\)) & 23.0 & 34.3 \\
        S-DETR & QD & SK (\(Q \cap S\)) & 38.3 & 57.5 \\ %
    \bottomrule
    \end{tabular}
    \label{t:cross_modal}
\end{table}

\begin{figure*}
    \centering
    \begin{tabular}{l c c c c c}
    \toprule
        \multirow{2}{*}{\rotatebox[origin=c]{90}{\parbox{3cm}{%
\centering \textbf{Sketchy}}}} & \begin{tabular}{@{\hskip 0pt}c@{\hskip 3pt} c@{\hskip 0pt}}
            \fbox{\includegraphics[width=0.05\textwidth,valign=m]{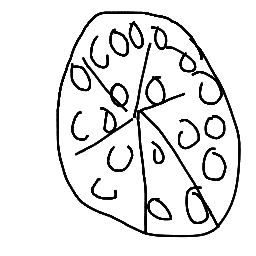}} & pizza
        \end{tabular} 
        & 
        \begin{tabular}{@{\hskip 0pt}c@{\hskip 3pt} c@{\hskip 0pt}}
            \fbox{\includegraphics[width=0.05\textwidth,valign=m]{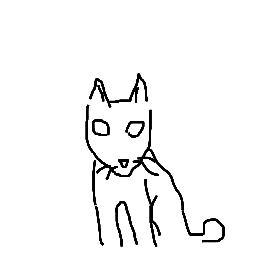}} & cat 
        \end{tabular} 
        & 
        \begin{tabular}{@{\hskip 0pt}c@{\hskip 3pt} c@{\hskip 0pt}}
            \fbox{\includegraphics[width=0.05\textwidth,valign=m]{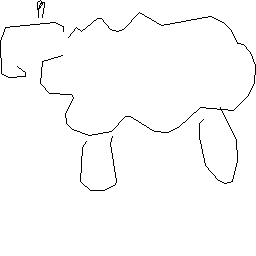}} & sheep 
        \end{tabular}& 
        \begin{tabular}{@{\hskip 0pt}c@{\hskip 3pt} c@{\hskip 0pt}}
            \fbox{\includegraphics[width=0.05\textwidth,valign=m]{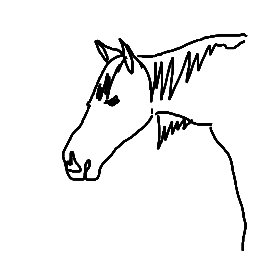}} & horse 
        \end{tabular}
        & 
        \begin{tabular}{@{\hskip 0pt}c@{\hskip 3pt} c@{\hskip 0pt}}
            \fbox{\includegraphics[width=0.05\textwidth,valign=m]{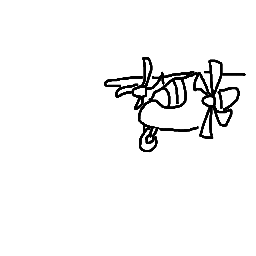}} & plane 
        \end{tabular}\\
        \addlinespace
        &
        \includegraphics[width=0.16\textwidth,valign=c]{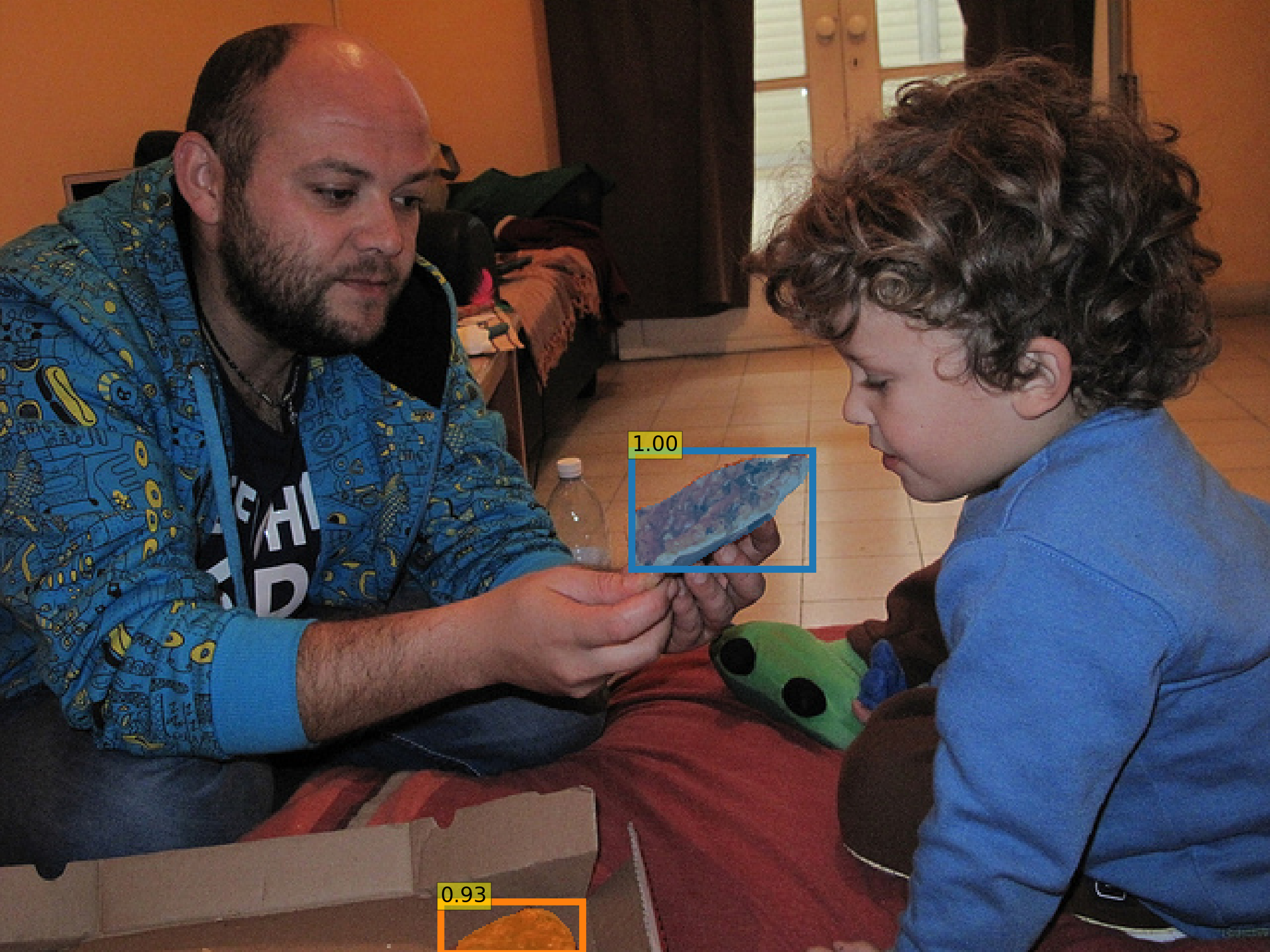} & \includegraphics[width=0.16\textwidth,valign=c]{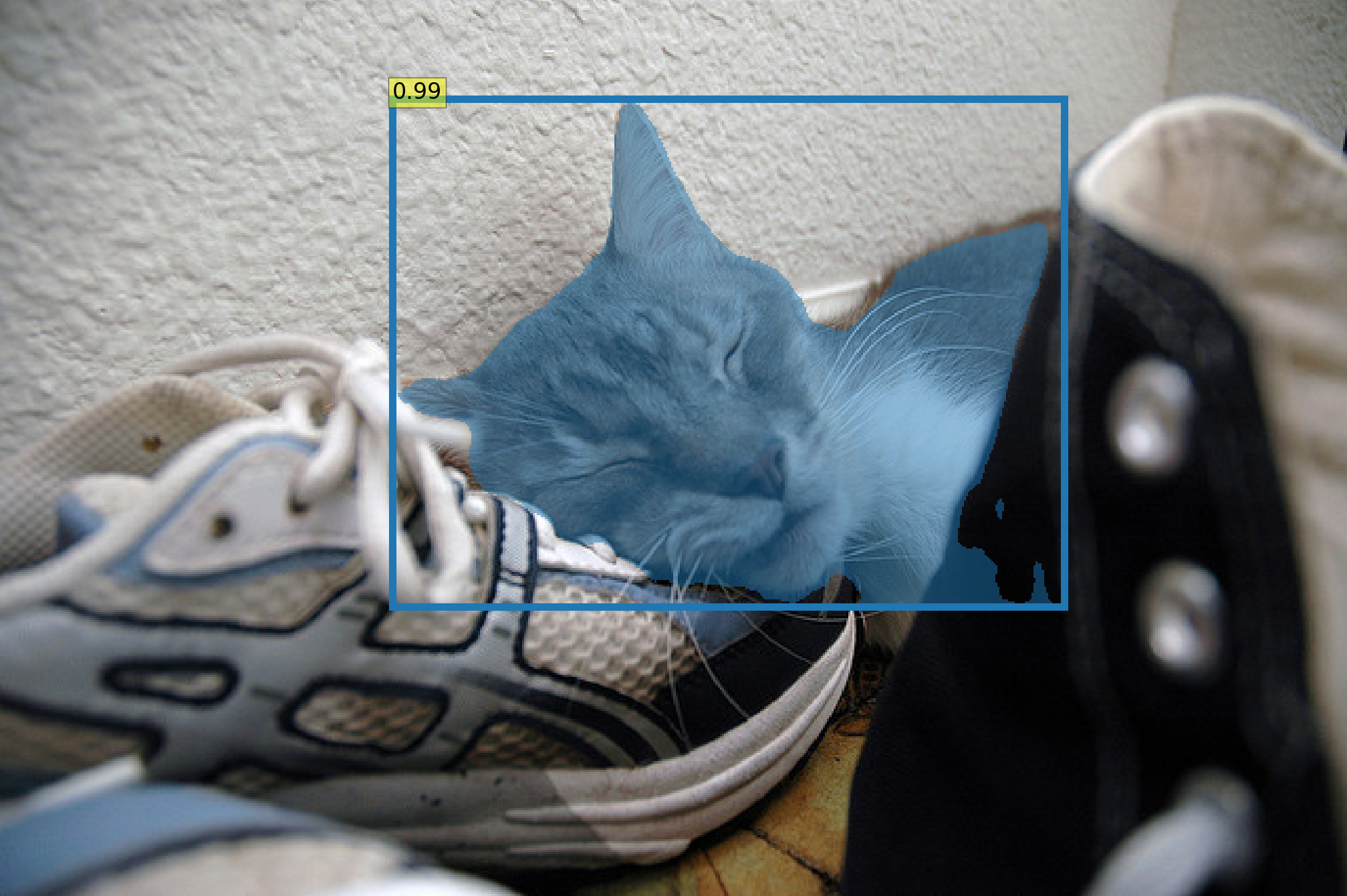} & \includegraphics[width=0.1\textwidth,valign=c]{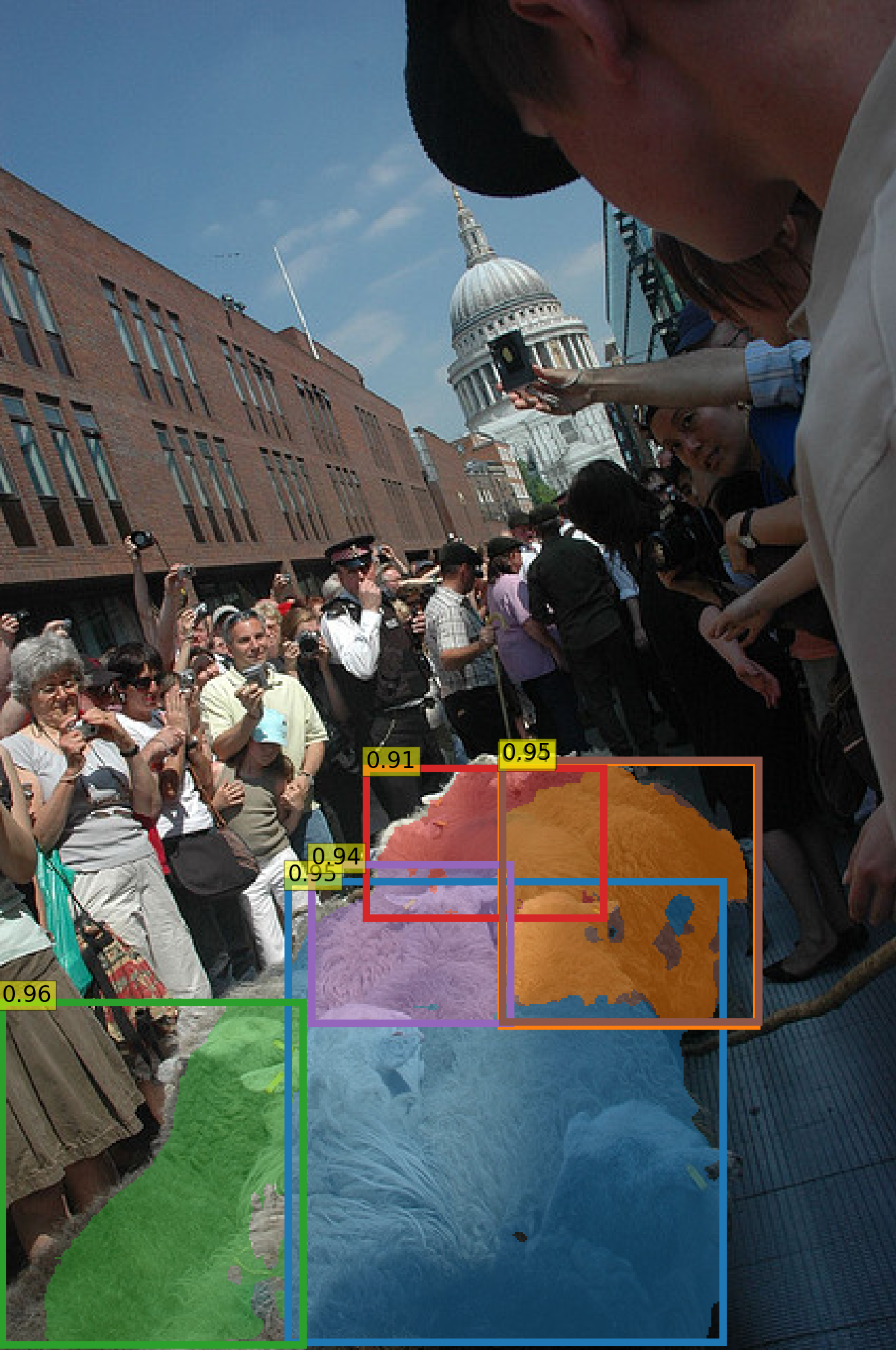} &  \includegraphics[width=0.16\textwidth,valign=m]{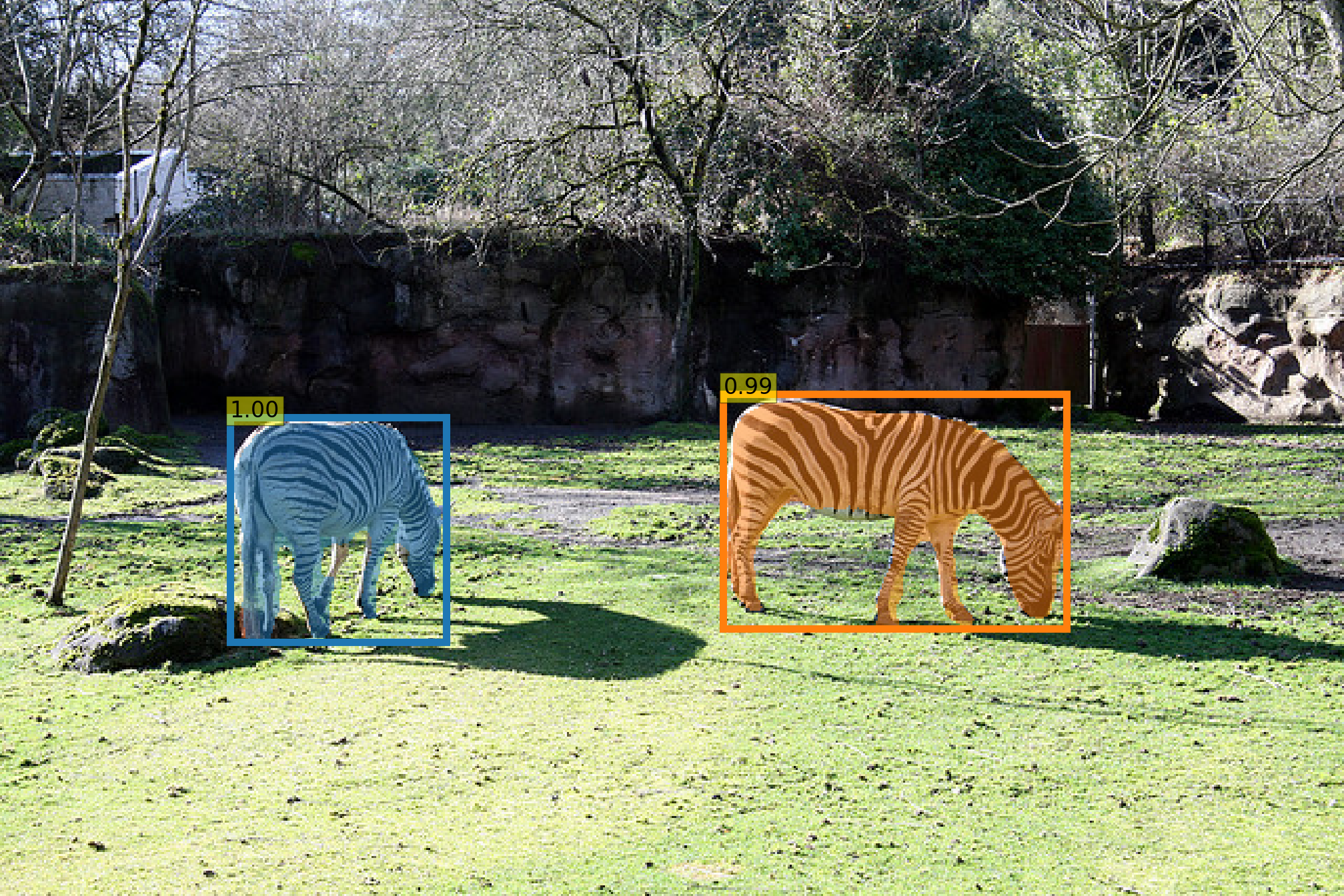}
        &  \includegraphics[width=0.16\textwidth,valign=m]{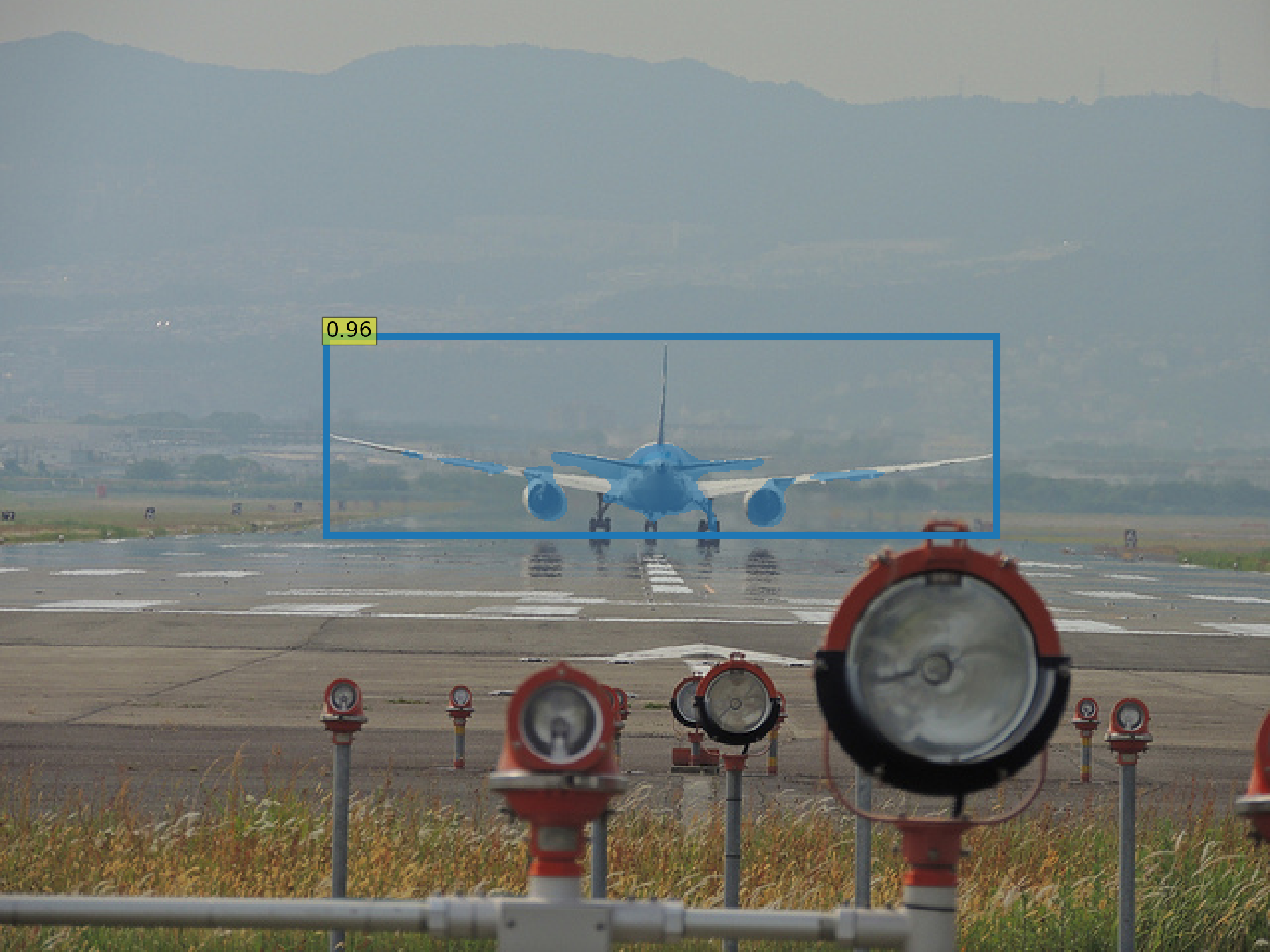}\\
        \midrule
        \multirow{2}{*}{\rotatebox[origin=c]{90}{\parbox{3cm}{%
\centering \textbf{QuickDraw}}}} & \begin{tabular}{@{\hskip 0pt}c@{\hskip 3pt} c@{\hskip 0pt}}
            \fbox{\includegraphics[width=0.05\textwidth,valign=m]{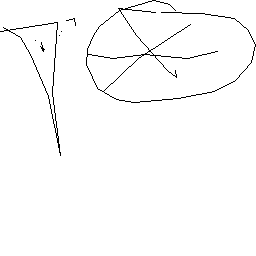}} & pizza
        \end{tabular} 
        & 
        \begin{tabular}{@{\hskip 0pt}c@{\hskip 3pt} c@{\hskip 0pt}}
            \fbox{\includegraphics[width=0.05\textwidth,valign=m]{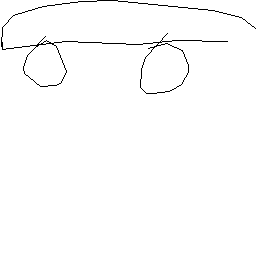}} & skate 
        \end{tabular} 
        & 
        \begin{tabular}{@{\hskip 0pt}c@{\hskip 3pt} c@{\hskip 0pt}}
            \fbox{\includegraphics[width=0.05\textwidth,valign=m]{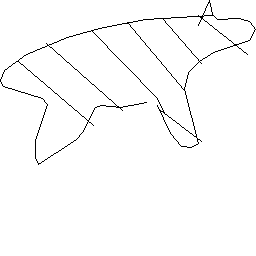}} & zebra 
        \end{tabular} & 
        \begin{tabular}{@{\hskip 0pt}c@{\hskip 3pt} c@{\hskip 0pt}}
            \fbox{\includegraphics[width=0.05\textwidth,valign=m]{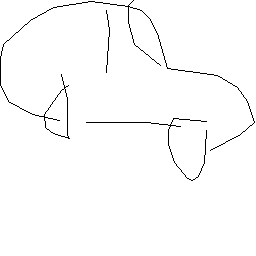}} & car 
        \end{tabular} & 
        \begin{tabular}{@{\hskip 0pt}c@{\hskip 3pt} c@{\hskip 0pt}}
            \fbox{\includegraphics[width=0.05\textwidth,valign=m]{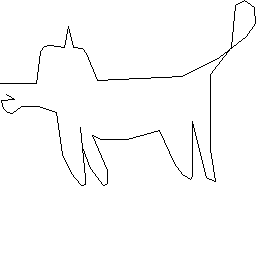}} & dog 
        \end{tabular}\\
        \addlinespace
        &
        \includegraphics[width=0.12\textwidth,valign=c]{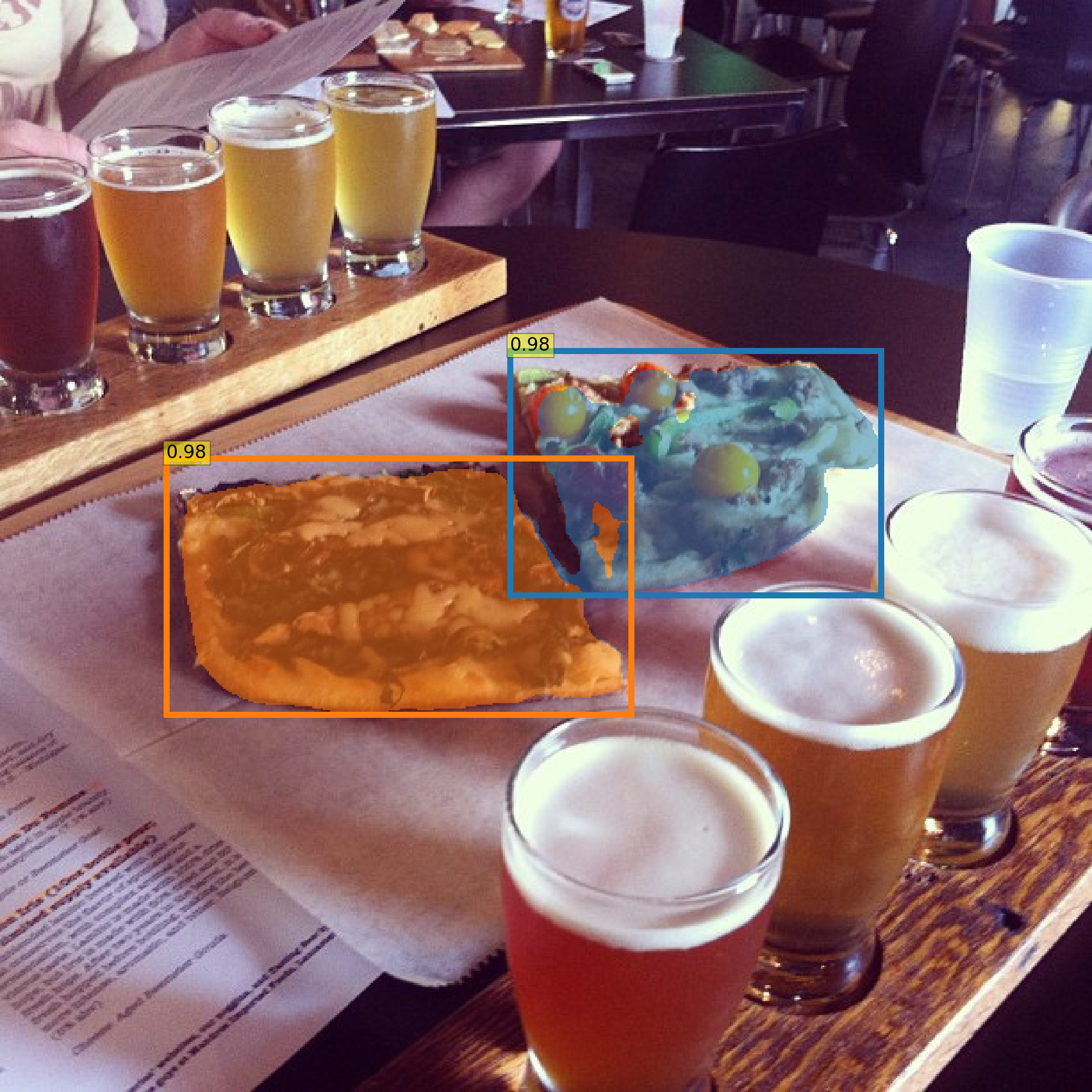} & \includegraphics[width=0.16\textwidth,valign=c]{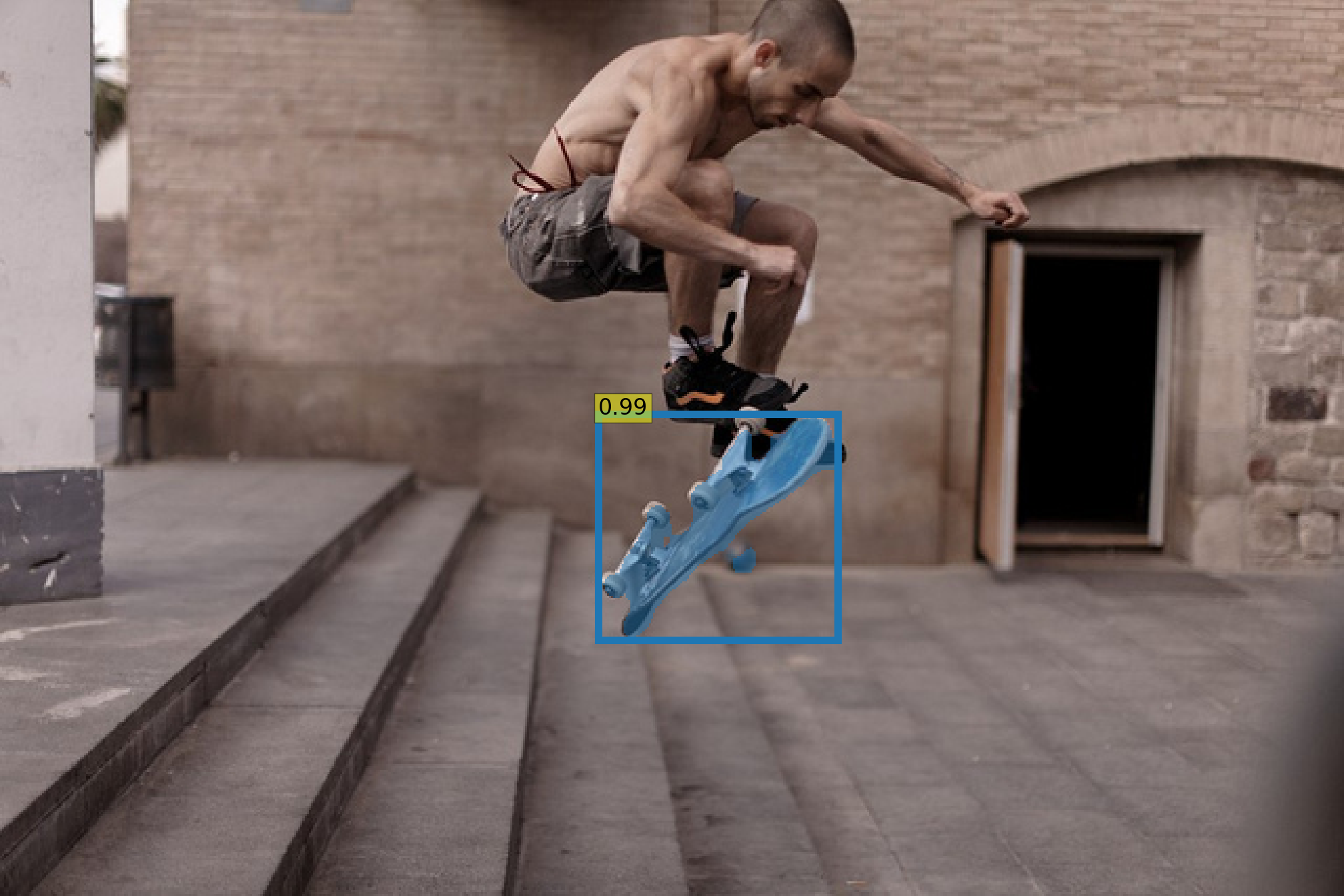} & \includegraphics[width=0.155\textwidth,valign=c]{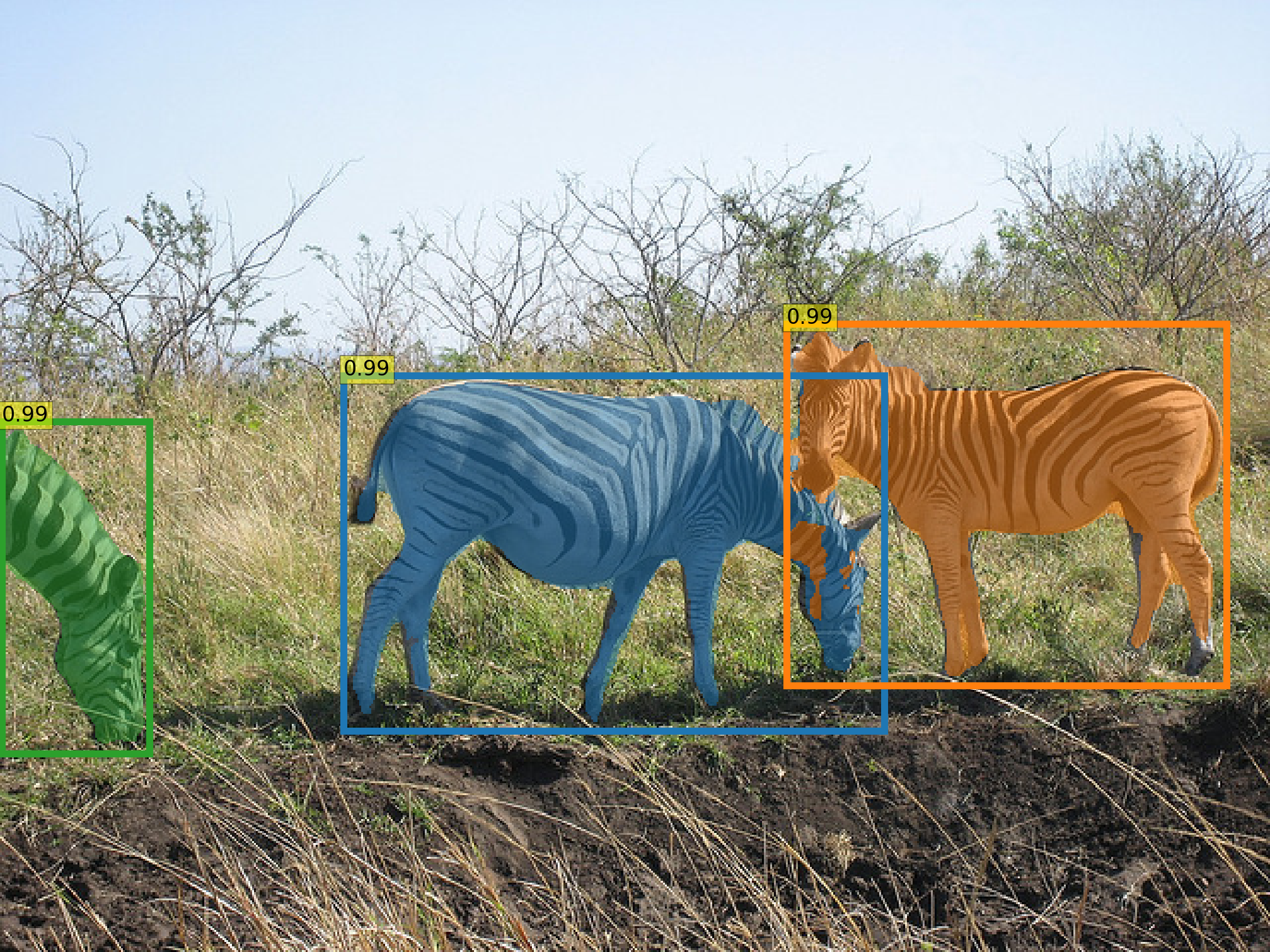} & \includegraphics[width=0.16\textwidth,valign=c]{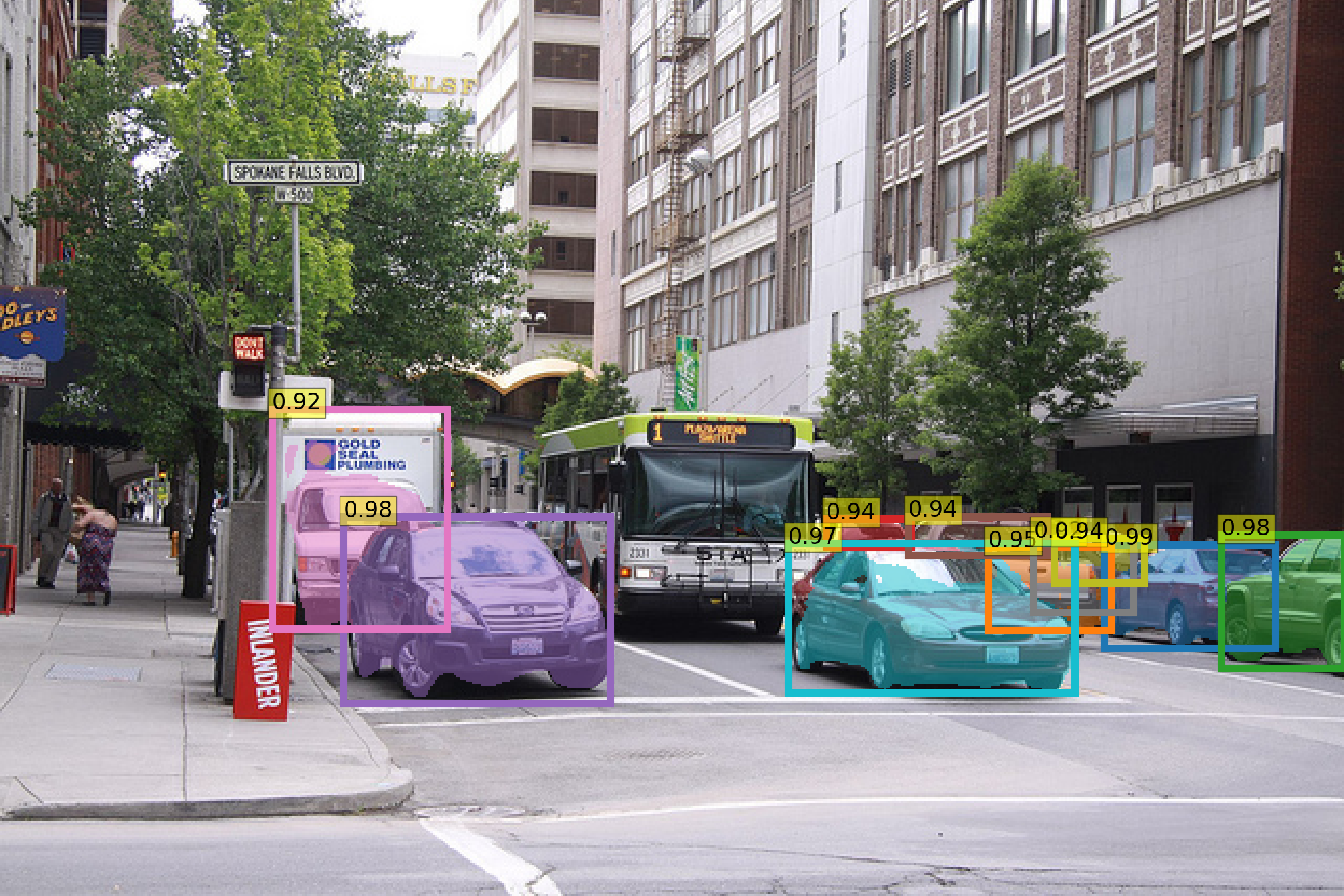} & \includegraphics[width=0.15\textwidth,valign=c]{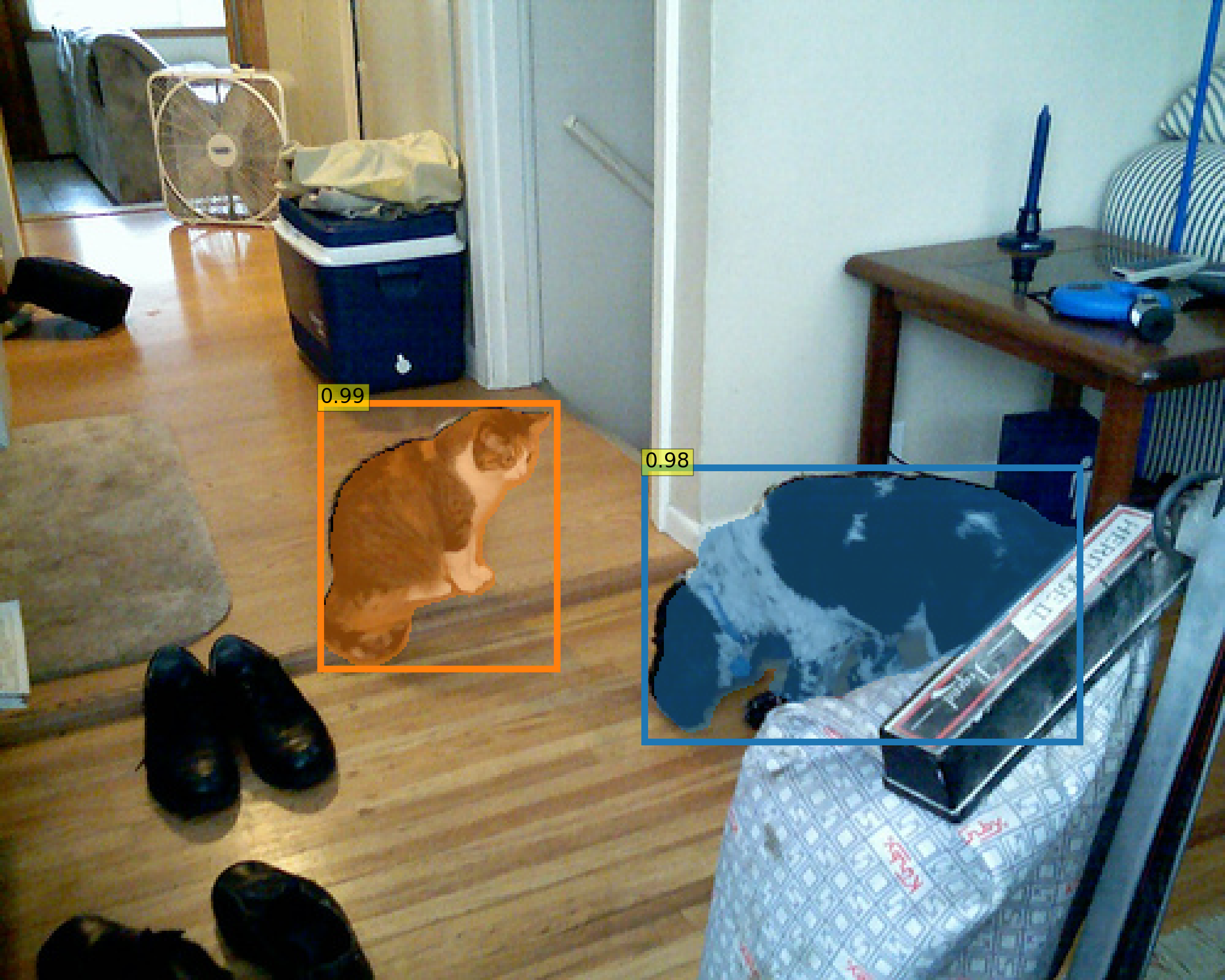} \\
        \bottomrule
        \addlinespace
    \end{tabular}
    \caption{SGOL qualitative evaluation for Sketchy (top) and QuickDraw (bottom). Each image depicts the detected bounding boxes and segmentation masks given a score higher than \(0.9\).}
    \label{f:qualitative}
\end{figure*}

\mysubtitle{Cross-dataset Evaluation:} 
We consider that moving to a whole disjoint setting is still not viable. Thus, we propose an intermediate stage according to a cross-dataset evaluation. At this point we consider a setting where we train on a given sketch dataset and we evaluate in another one. It can be noticed that this evaluation deals with two challenges of the sketch domain. First, there is a huge gap on the level of abstraction between the training and testing sketches. Second, some new sketch classes are also evaluated. 
Note that we do not consider our approach to be zero-shot as these classes have been already seen by our pre-trained object detector as images. 
We report the evaluation on only the classes that are common among to both datasets (\(Q \cap S\)) or that are disjoint (\(Q \setminus S\)). We do not consider the case \(S\setminus Q\) as it only contains a single class. Table~\ref{t:cross_modal} provides the cross-modal evaluation results.

The first conclusion we draw is that the proposed model is always better than tough-to-beat baseline on both QuickDraw and Sketchy, showing a promise on the cross-domain performance. In other words, since the sketch classifier can not properly classify the different domain sketches, the baseline numbers are quite low which is the limitation we expressed previously. 

Moreover, we notice a $10$ point drop on both AP and mAP for Sketch-DETR compared to the same evaluation setting. While this drop is significantly magnified for T2B to 30 or 20 points depending on the dataset. Thus, it can be stated that Sketch-DETR is more robust than T2B baseline in terms of the performance drop compared to Table~\ref{t:detection_comparison}. It lies in the fact that utilizing feature space in Sketch-DETR leads to a smoother space between sketches and images. 

Finally, the results demonstrate that there is a challenging intermediate step that should be properly addressed before facing a more arduous task of zero-shot. We clearly see that current models can not handle the unseen sketches, which calls for a unique architecture for this specific setting.

\subsection{Qualitative results}
 
Figure~\ref{f:qualitative} shows some qualitative examples for both sketch datasets. It shows the %
behavior of our model in both tasks of SGOL, namely, object detection and instance segmentation. The sketches from the two different datasets depicts the difference in the degree of abstraction of the queried concepts. This gap of depiction can be observed for instance, in the sketches belonging to `pizza' and `horse'/`zebra'. 

The proposed model is not only able to understand and perfectly localize examples such as `zebra' where the sketch is highly abstract, but also interprets the stripes that help humans to understand that this is, indeed, a `zebra'. Similarly, in both the cases of `pizza' in multi-object scenes the instances have been perfectly delineated. Moreover, our model is able to detect a `plane' although it has a different pose.

Even in the failed cases, where two classes are semantically or visually similar, some interesting conclusions can be observed. In particular, in the queries `horse', `car' and `dog', whose sketches are easily confused even by human observers, the model picks instances with many visual shape similarity. These examples are easily confused with that of `zebra', `truck' and `cat' respectively. Handling cluttered scenes with several occlusions becomes very challenging while obtaining instance level pixel labels (see `sheep').

\section{Conclusion}\label{s:conclusion}
This work has presented an in-depth study of the SGOL task. Even though previous works have focused on the sketch-guided object detection, we believe that such a new task requires a rigorous analysis of the performance of straightforward but efficient methods. 
With this aim, we put forward a tough-to-beat baseline that is able to outperform the state-of-the-art. 
Then, we have proposed a novel model which is inspired by the DETR object detection, that is able to deal with unconstrained scenarios. Moreover, these two models have been naturally extended to the novel task of SGOL at instance segmentation level. We experimentally show that our proposed frameworks overcome the recent state-of-the-art approaches. 
Currently, SGOL is evaluated at category-level, however, we firmly believe that a long term research line should move towards a visual similarity in terms of pose, shape, etc.

\clearpage 

{\small
\bibliographystyle{ieee_fullname}
\bibliography{references}
}

\end{document}